\documentclass{article}

\usepackage{microtype}
\usepackage{graphicx}
\usepackage{caption}
\usepackage{subcaption}

\usepackage{booktabs} 
\usepackage[draft]{hyperref}
\usepackage{amsmath}

\usepackage[accepted]{whi2020}
\icmltitlerunning{Towards Ground Truth Explainability for Tabular Data}

\begin{document}

\twocolumn[
\icmltitle{Towards Ground Truth Explainability on Tabular Data}



\icmlsetsymbol{equal}{*}

\begin{icmlauthorlist}
\icmlauthor{Brian Barr}{c1ny}
\icmlauthor{Ke Xu}{nyu}
\icmlauthor{Claudio Silva}{nyu,nyu2}
\icmlauthor{Enrico Bertini}{nyu}
\icmlauthor{Robert Reilly}{c1tx}
\icmlauthor{C. Bayan Bruss}{c1va}
\icmlauthor{Jason D. Wittenbach}{c1ma}
\end{icmlauthorlist}

\icmlaffiliation{c1ny}{Center for Machine Learning, Capital One, New York, NY}
\icmlaffiliation{c1va}{Center for Machine Learning, Capital One, Maclean, VA}
\icmlaffiliation{c1ma}{Center for Machine Learning, Capital One, Cambridge, MA}
\icmlaffiliation{c1tx}{Card Machine Learning and Technology, Capital One, Plano, TX}

\icmlaffiliation{nyu}{Tandon School of Engineering, New York University, NY}
\icmlaffiliation{nyu2}{Center for Data Science, New York University, NY}

\icmlcorrespondingauthor{Brian Barr}{brian.barr@capitalone.com}
\icmlcorrespondingauthor{Claudio Silva}{csilva@nyu.edu}

\icmlkeywords{Machine Learning, ICML}

\vskip 0.3in
]



\printAffiliationsAndNotice{}  

\begin{abstract}
In data science, there is a long history of using synthetic data for method development, feature selection and feature engineering. Our current interest in synthetic data comes from recent work in explainability.  Today's datasets are typically larger and more complex - requiring less interpretable models.  In the setting of \textit{post hoc} explainability, there is no ground truth for explanations.  Inspired by recent work in explaining image classifiers that does provide ground truth,  we propose a similar solution for tabular data.  Using copulas, a concise specification of the desired statistical properties of a dataset, users can build intuition around explainability using controlled data sets and experimentation. The current capabilities are demonstrated on three use cases: one dimensional logistic regression, impact of correlation from informative features, impact of correlation from redundant variables. 

\end{abstract}

\begin{figure*}
    \centering
       \begin{subfigure}[b]{0.24\textwidth}
       \includegraphics[scale=0.2]{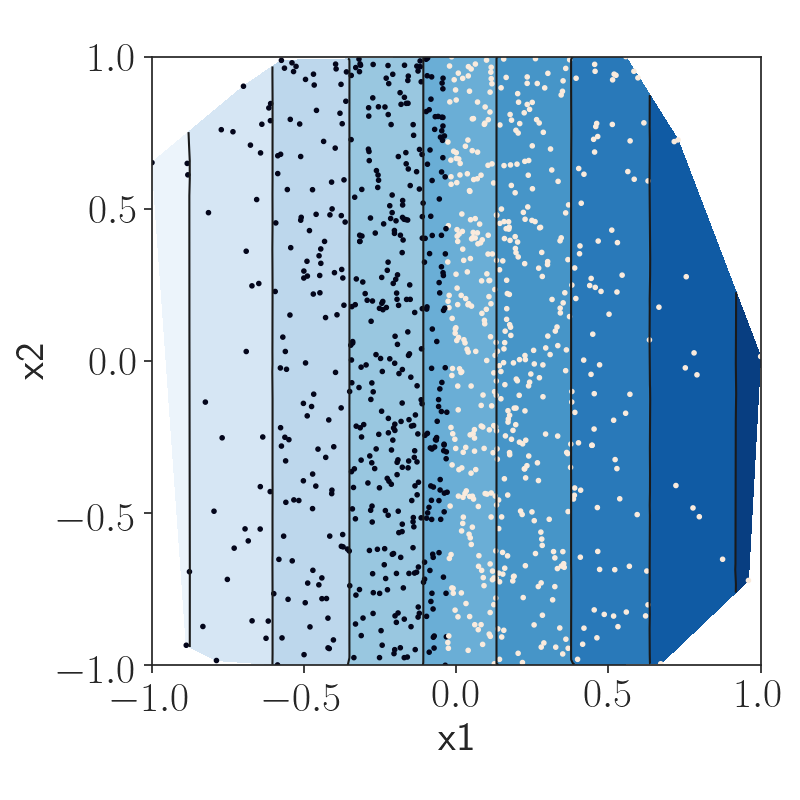}
       \caption{}
       \label{fig:1} 
    \end{subfigure}
    \begin{subfigure}[b]{0.24\textwidth}
       \includegraphics[scale=0.2]{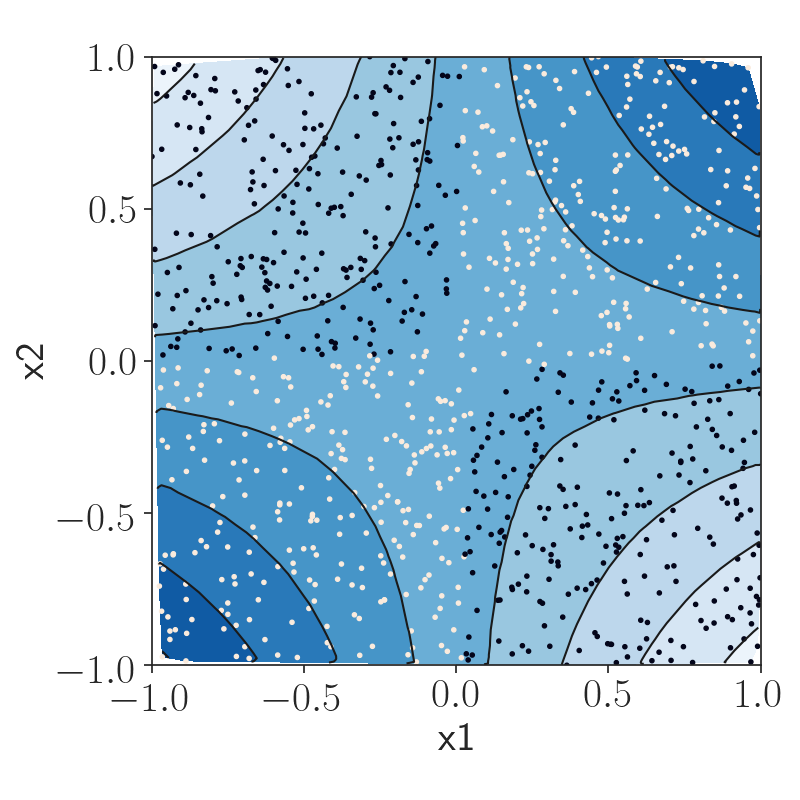}
       \caption{}
       \label{fig:2}
    \end{subfigure}
    \begin{subfigure}[b]{0.24\textwidth}
       \includegraphics[scale=0.2]{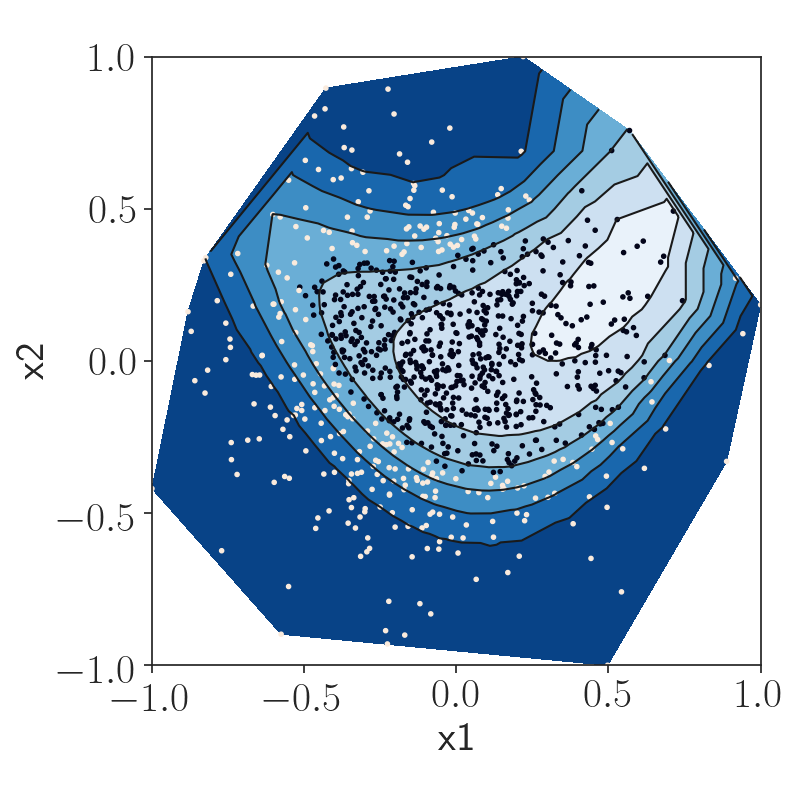}
       \caption{}
       \label{fig:3}
    \end{subfigure}
    \begin{subfigure}[b]{0.24\textwidth}
       \includegraphics[scale=0.2]{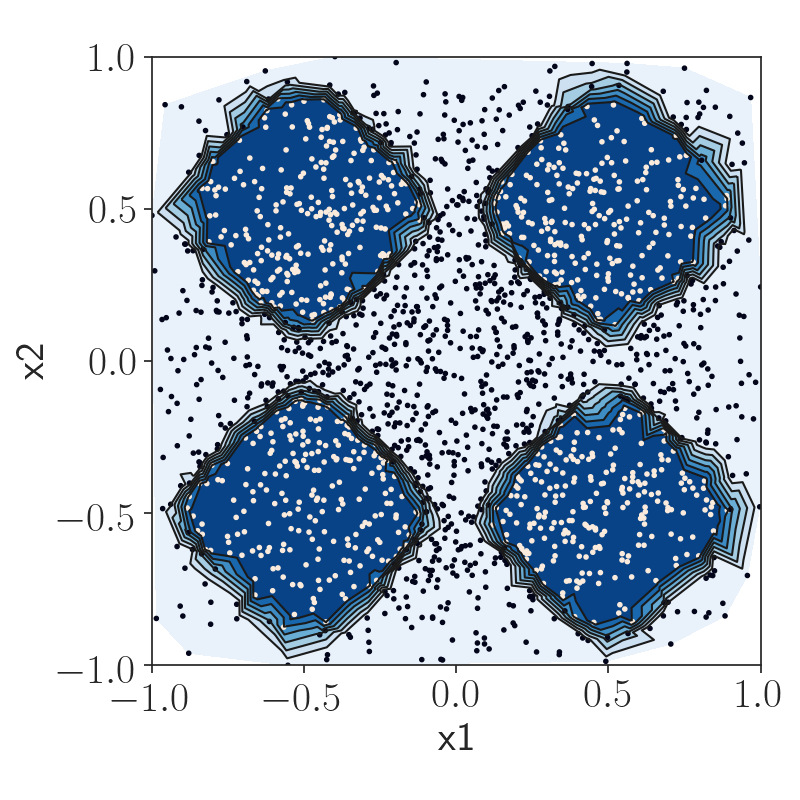}
       \caption{}
       \label{fig:4}
    \end{subfigure}
    \caption{Sample datasets - (a) linear, (b) nonlinear, (c) rosenbrock function, and (d) rastrigin function.  The background has contours of probability with a scatter plot of the sampled points, colored by class. The datasets in (a) and (b) will be used later for experiments in Section 4.}
\end{figure*}

\section{Introduction}
The combination of large public datasets and novel machine learning architectures has provided state of the art predictive power in many diverse fields, such as computer vision and machine translation.  These models are largely regarded as opaque \textit{black-box} models.  With their prevalence and increasing adoption, an active field of research is eXplainable Artificial Intelligence (XAI), which has sought to provide explanations for their predictions.

One avenue of research is on building interpretability into the model architecture~\cite{kim2015mind, gupta2016lattices, lee2019locallylinear}.   We focus on the area \textit{post hoc} explanations -- which occurs after model training.  Currently there is no one size fits all solution.  The method of explanation depends on model type~\cite{chen2018looks, krakovna2016increasing, grath2018interpretable}, desired granularity~\cite{ribiero2016lime,ibrahim2019gam, dhur2018explanations, bhatt2020evaluating,linden2019global}, and audience~\cite{wachter2017counterfactual,bhatt2020explainable}.

In support of the methods, there are a growing number of packages that seek to provide an umbrella of methods such as AIX360~\cite{aix360-sept-2019}, ELI5~\cite{eli5}, and Alibi~\cite{alibi}.

Early methods were focused on explaining image classifiers.  The use of sensitivities of the output class based on the input image pixels, provides a visual and immediately interpretable explanation for the classifier's prediction.  For tabular data, the intuitive visual nature of sensitivities is not a natural metaphor. Additionally, where as computer vision typically relies on correlations between pixels as features, for tabular data that can be detrimental~\cite{aas2019explaining}.


An ongoing challenge in XAI is the lack of ground truth. To add to the landscape of XAI, and move towards ground truth explainability for tabular data, we provide a flexible synthetic data generation method allowing generation of arbitrarily complex data.  The current implementation is focused on the task of binary classification.

The paper is structured as follows: previous work is discussed in Section 2, Section 3 presents the method used to construct the synthetic tabular data, and Section 4 shows some results from three use cases: one dimensional logistic regression,  impact of correlation from informative features, and impact of correlation from redundant variables.

\section{Previous work}
Early use cases of synthetic data focused on the tasks of feature and model selection~\cite{guyonTechReport}. This method is available in the scikit-learn~\cite{scikit-learn} module $make\_classification$.  An alternative method of generating tabular data for classification is to generate clusters and apply class labels to them.

Another approach is to model joint probability $P(\boldsymbol{X}, y)$ from an actual dataset. This can be helpful in dealing with sensitive data and as an aid in sharing data where there are privacy and regulatory restrictions on the use of actual data~\cite{ping2017datasynthesizer, howe2017synthetic, goncalves2020medical}.  Techniques used range from probabilistic models, to Bayesian networks, to  generative adversarial neural networks (GANS).  In finance, it is typical to use copulas.  The theory of copulas  has been developed considerably in mathematics, statistics, actuarial science, with significant interest in their application to finance \cite{genest2009advent}, and their  misuse may have led the financial crisis \cite{salmon2012formula}.  However, methods that mimic the statistics of other datasets are incapable of providing ground truth for explanations - since they lack the complete data generation process that imposes a functional dependence between $\boldsymbol{X}$ and $y$.


Our research is inspired from recent work in interpreting the image classifiers trained with a carefully crafted dataset that controls the relative feature importance \cite{yangBenchmarkingAttributionMethods2019}. In this case, the model can be quantitatively evaluated in the form of known foreground and background images by providing ground truth of local feature importances.

We propose a similar method for tabular data. We use copulas to define the correlation structure and marginal distributions of the independent features.  We specify the dependence of the underlying probability field as a symbolic expression.  Binary class labels are assigned by setting a threshold probability.  This method provides global explanations since we prescribe the coefficients of the terms in the symbolic expression. In some instances, where we build models only from informative features,  we can provide ground truth local attributions. 

Our contributions are providing a unique and flexible method for synthetic tabular data generation suitable for current model architectures and demonstration of its use in informative experiments highlighting that not all correlation in inputs change local attributions.

\begin{figure*}
    \centering
    \begin{subfigure}[b]{0.5\columnwidth}
        \includegraphics[width=\textwidth]{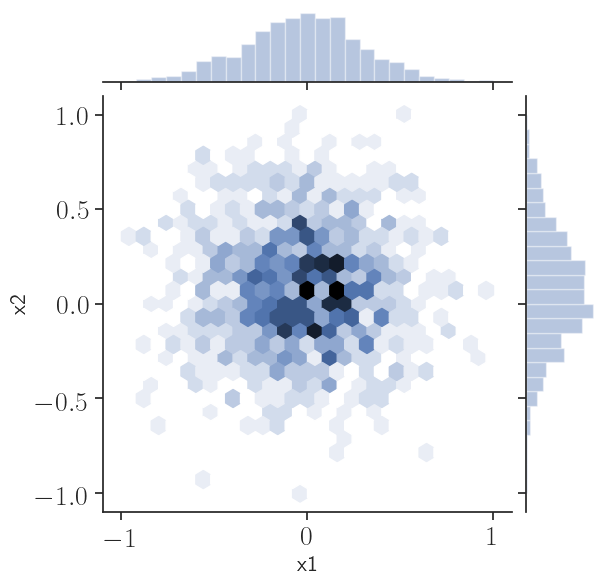}
        \caption{}
        \label{fig:joint_logistic}
    \end{subfigure}
    \begin{subfigure}[b]{0.5\columnwidth}
        \includegraphics[width=\textwidth]{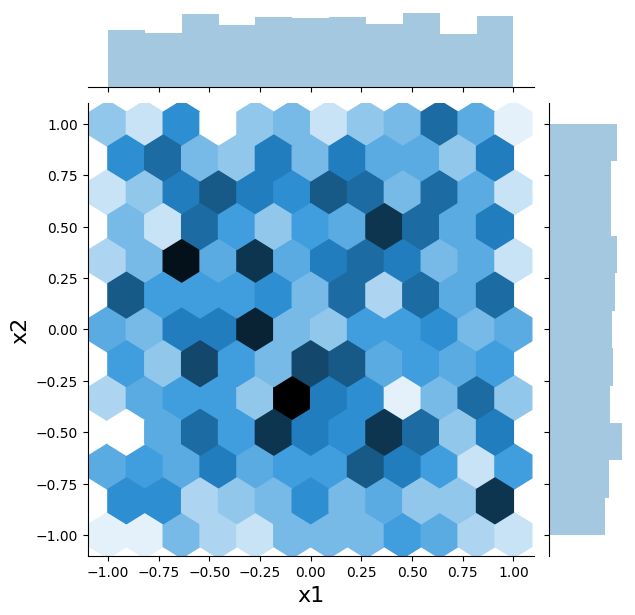}
        \caption{}
        \label{fig:joint_2d_no_cov} 
    \end{subfigure}
    \begin{subfigure}[b]{0.5\columnwidth}
       \includegraphics[width=\textwidth]{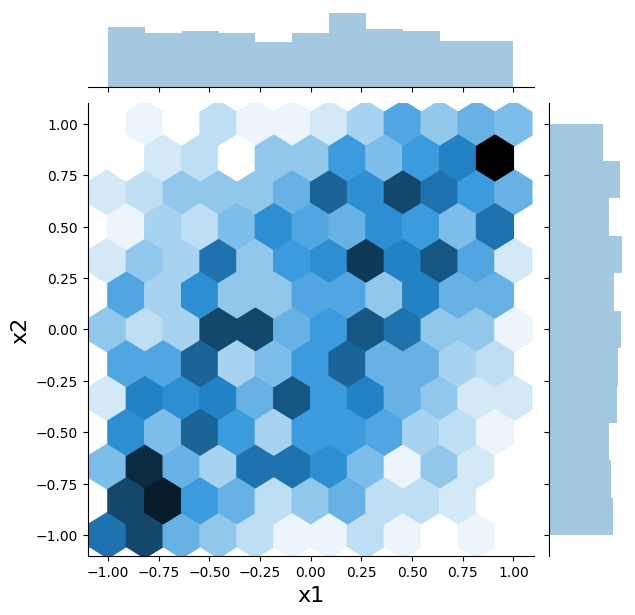}
       \caption{}
       \label{fig:joint_2d_cov}
    \end{subfigure}
    \caption{Joint probability plot for features $x_1$ and $x_2$ (a) with Gaussian marginals (b) with uniform marginals with no correlation and (c) uniform marginals with positive correlation.}
\end{figure*}


\section{Synthetic data generation}

The generation of synthetic data seeks a method to provide the joint probability $P(\boldsymbol{X}, y)$ where $\boldsymbol{X}$ are the input features, and $y$ is the output variable, and to draw samples from that joint distribution.  From those samples, we will fit machine learned models that approximate the conditional probability $P(y|X)$ (via possibly black box models) and to provide explanations for those models.  

We separate  the input feature vectors, \(\boldsymbol{X}\),  into three categories - informative, redundant, and nuisance features.

\begin{equation}
 \boldsymbol{X} = \left( \boldsymbol{X}_{I} | \boldsymbol{X}_{r} | \boldsymbol{X}_{n}\right)
\end{equation}

Informative features, $\boldsymbol{X}_{I}$, used to determine the binary labels, are specified by a copula.  A multivariate distribution can be separated into a set of marginal distributions and the correlation structure between them. Copula theory is the mathematical framework of the separation of the correlation structure from the marginal distributions of the feature vectors. See \cite{jaworski_2013, nelsen_1999} for further details.  The current library supports any marginal distribution available in \textit{scipy.stats}.  The results for this paper use a multivariate Gaussian copula. 

Redundant features, $\boldsymbol{X}_{r}$, are a random linear combination of informative features.  Nuisance features, $\boldsymbol{X}_{n}$ are random uncorrelated vectors drawn from the interval [-1, 1] which are useful as benchmarks to set the lower bound on explainability.  Being purely random and not contributing to the specification of the labels, any feature found to have lower importance than a nuisance feature should be subjected to further scrutiny and possibly removed from the model.

The final step in the process is to generate binary labels for the inputs.  First, a scalar regression value is created from the informative features via a symbolic expression using the \textit{sympy} python module.  
\begin{equation}
 \vec{y}_{reg} = f\left( \boldsymbol{X}_{I} \right)
\end{equation}
\begin{equation}
 h_ k (y_{reg}) =  \frac{\mathrm{1} }{\mathrm{1} + e^{- k(y_{reg} - y_0)}}
\end{equation}

To generate binary classification probabilities, the regression value is squashed by a sigmoid to the range [0,1].  After setting a threshold probability, class labels are determined.

Additional post processing allows the addition of Gaussian noise to the informative features.  A random seed can be specified so that repeated runs of a synthetic dataset yield the same values. 

\section{Experiments}
This section demonstrates some of the capabilities of the synthetic tabular data through the process of modeling and providing local attributions via the SHAP library~\cite{lundbergUnifiedApproachInterpreting}.

\subsection{Logistic regression}
The first synthetic data set is for two features,  $x_1, x_2 \in [-1,1]$, with no covariance and Gaussian marginal distributions for both.  The joint probability plot is shown in Figure~\ref{fig:joint_logistic}.

The symbolic regression expression is $y = x_1$. The probability values and class labels are shown in Figure~\ref{fig:1}.  One thousand samples are generated.

\begin{figure}[ht]
    \centering
    \includegraphics[width=\columnwidth]{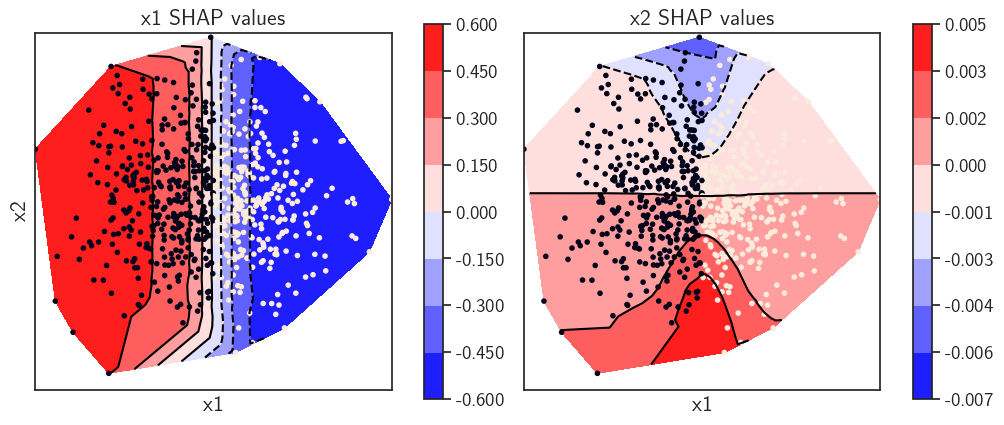}
    \caption{Contours of SHAP values for $x_1$ (left) and $x_2$ (right) for the simple 1-D logistic regression model.}
    \label{fig:logistic_shap_vals}
\end{figure}

The dataset is split 70/30 into a training and test set, with a logistic regression model fit to the training data. The AUC of the model is 99.8\% and the coefficients of the model are $[11.98,  0.04]$. To provide local attributions, we fit a SHAP KernelExplainer to the training set with the results shown in Figure~\ref{fig:logistic_shap_vals}.

The SHAP values for $x_1$ dominate by two orders of magnitude, roughly in keeping with the relative global importance found in the coefficients.  Strong left to right symmetry is broken only in sparsely populated regions.  Interesting to note that the SHAP values for $x_2$ also display symmetry from top to bottom - with the opposite sign of the coefficient.

\begin{figure*}[htbp]
    \centering
    \includegraphics[width=0.8\textwidth]{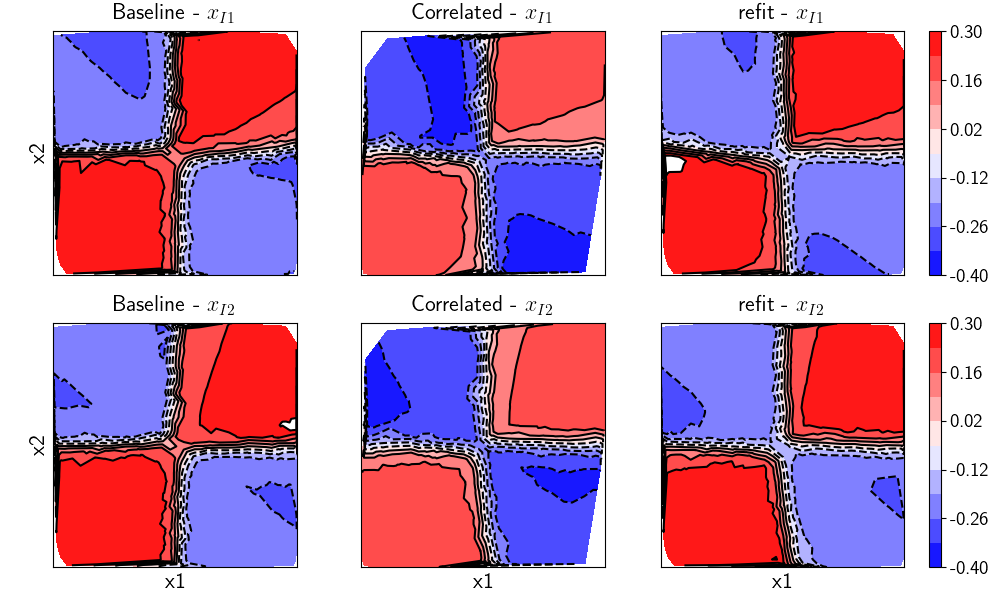}
  
    \caption{Contours of SHAP values for $x_1$ (top row) and $x_2$ (bottom row) for the baseline data and model (left), correlated model and data (center), and correlated model fit with baseline data (right).}
    \label{fig:2D_shap_vals}
\end{figure*}
\subsection{Presence of correlation in informative features}
We investigate the impact of correlation between the input features on model explanations.  It is well known that the presence of correlation can alter a model's explanations, see for example~\cite{breiman2001,aas2019explaining}.  


We again generate datasets for two features,  $x_1, x_2 \in [-1,1]$; the first with uniform marginal distributions, a baseline dataset with no correlation whose joint distribution can be seen in Figure~\ref{fig:joint_2d_no_cov}, and a second dataset, see Figure~\ref{fig:joint_2d_cov} with the covariance specified as:

\[ cov = \left[ \begin{array}{cc}
1.0 & 0.5  \\
0.5 & 1.0   \end{array} \right].\]

The symbolic regression expression for this experiment is:
\[ y_{reg} = cos(x_1^ 2 \cdot \pi / 180) - sin(x_2 \cdot \pi / 180) + x_1 \cdot x_2 \]

We hold the probability field constant between the datasets. The  probability values and class labels are shown in Figure~\ref{fig:2}. 

The dataset is split 70/30 into a training and test set, with a dense network with three hidden layers and a total of 255 weights in Tensorflow  fit to the training data. The AUC of the resulting model is 100\%. To provide local attributions, we fit a SHAP DeepExplainer to the training set with the results shown in Figure~\ref{fig:2D_shap_vals}.

There is an apparent decrease in SHAP values in the presence of correlation in the input features.  Recall that SHAP values are relative with respect to the expectation of the output.  With correlation, their density is drawn out of quadrants 2 and 4, and placed in quadrants 1 and 3, leading to a higher expected predicted probability due to the imbalance of the class labels.  If we account for that effect by refitting the explainer with the baseline data (which does not suffer the same level of imbalance), the SHAP values look essentially like those from the uncorrelated inputs (as shown in the last column of Figure~\ref{fig:2D_shap_vals})

In this context, the apparent effect of correlation is an artifact  induced from the expectation of the correlated model predictions over the correlated training data -- it is not a bias that the model has learned during training.

\subsection{Presence of redundant variables}
In contrast to the experiment in the previous section, this section considers the effect of correlation with unimportant features -- since the redundant features are not explicitly used in the symbolic expression used to derive the binary labels. We reuse the baseline from the previous section.  We create a second dataset by augmenting the baseline informative features with two redundant and two nuisance features.  



The joint distribution of the informative features, the symbolic expression that maps the informative features to the binary labels all remain unchanged.  We keep the same train/test split.  The only change to the dense network is to increase the dimension of the input layer to accommodate the two redundant and two nuisance features.

\begin{figure*}
    \centering
    \includegraphics[width=0.95\textwidth, keepaspectratio]{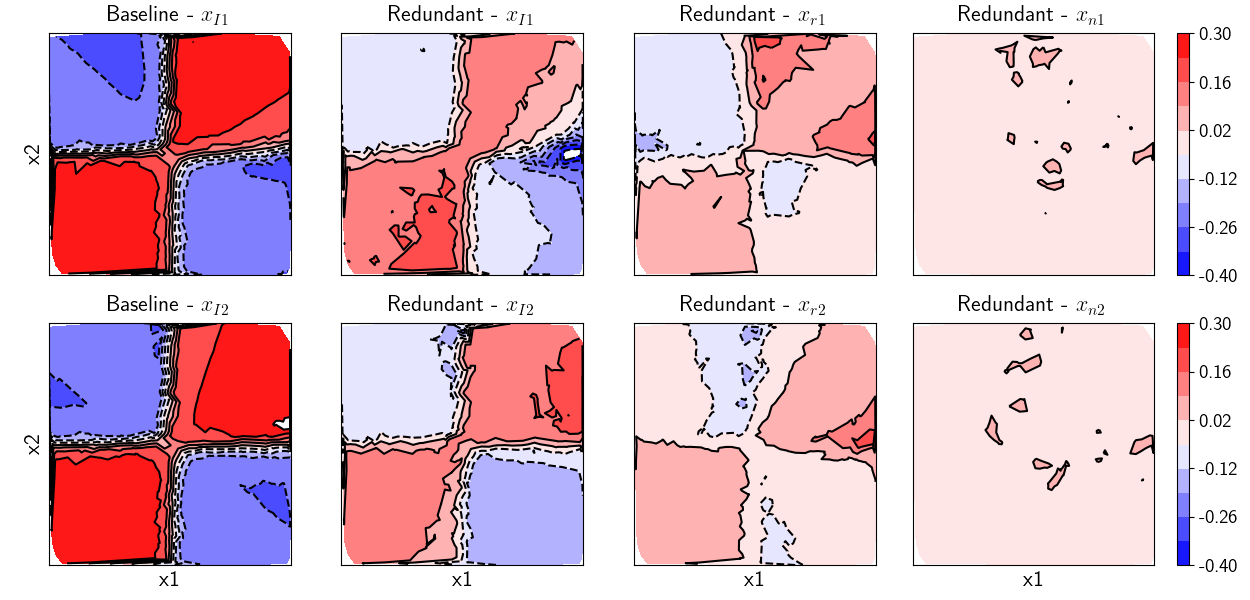}
    \caption{Contours of SHAP values for $x_1$ (top row) and $x_2$ (bottom row) for the baseline data and model(left), redundant feature model for three components of explanation: informative, redundant, and nuisance.}
    \label{fig:2D_shap_vals_with_redundant}
\end{figure*}

Redundant features have a significant impact on SHAP's perception of feature importance. Summing the individual components of explanations for the redundant feature model does not recover the values obtained from the baseline model.  


\section{Conclusions}
In this paper, we have described a method of generating synthetic tabular data utilizing copulas and symbolic expressions that provides users the ability to control the complexity of the resulting dataset. We have briefly demonstrated how the origin  of correlation amongst inputs (informative vs redundant) influences model explainability. This method is a powerful investigative tool for data scientists, model developers, and explainable AI method developers.

Future work includes: further experimentation with more complex data sets, bridging the gap between global explanations (provided by our symbolic expression) and local explanations, investigation of categorical and ordinal variables, as well as improving local attributions for tabular data.

At the time of publication, we were working on making the repository accessible.  Please contact a corresponding author for further details.





\bibliography{paper}

\begin{thebibliography}{29}
\providecommand{\natexlab}[1]{#1}
\providecommand{\url}[1]{\texttt{#1}}
\expandafter\ifx\csname urlstyle\endcsname\relax
  \providecommand{\doi}[1]{doi: #1}\else
  \providecommand{\doi}{doi: \begingroup \urlstyle{rm}\Url}\fi

\bibitem[Aas et~al.(2019)Aas, Jullum, and L{\o}land]{aas2019explaining}
Aas, K., Jullum, M., and L{\o}land, A.
\newblock Explaining individual predictions when features are dependent: More
  accurate approximations to shapley values.
\newblock \emph{CoRR}, abs/1903.10464, 2019.
\newblock URL \url{http://arxiv.org/abs/1903.10464}.

\bibitem[Arya et~al.(2019)Arya, Bellamy, Chen, Dhurandhar, Hind, Hoffman,
  Houde, Liao, Luss, Mojsilovi\'c, Mourad, Pedemonte, Raghavendra, Richards,
  Sattigeri, Shanmugam, Singh, Varshney, Wei, and Zhang]{aix360-sept-2019}
Arya, V., Bellamy, R. K.~E., Chen, P.-Y., Dhurandhar, A., Hind, M., Hoffman,
  S.~C., Houde, S., Liao, Q.~V., Luss, R., Mojsilovi\'c, A., Mourad, S.,
  Pedemonte, P., Raghavendra, R., Richards, J., Sattigeri, P., Shanmugam, K.,
  Singh, M., Varshney, K.~R., Wei, D., and Zhang, Y.
\newblock One explanation does not fit all: A toolkit and taxonomy of ai
  explainability techniques, September 2019.
\newblock URL \url{https://arxiv.org/abs/1909.03012}.

\bibitem[Bhatt et~al.(2020{\natexlab{a}})Bhatt, Weller, and
  Moura]{bhatt2020evaluating}
Bhatt, U., Weller, A., and Moura, J.~M.
\newblock Evaluating and aggregating feature-based model explanations.
\newblock \emph{arXiv preprint arXiv:2005.00631}, 2020{\natexlab{a}}.

\bibitem[Bhatt et~al.(2020{\natexlab{b}})Bhatt, Xiang, Sharma, Weller, Taly,
  Jia, Ghosh, Puri, Moura, and Eckersley]{bhatt2020explainable}
Bhatt, U., Xiang, A., Sharma, S., Weller, A., Taly, A., Jia, Y., Ghosh, J.,
  Puri, R., Moura, J.~M., and Eckersley, P.
\newblock Explainable machine learning in deployment.
\newblock In \emph{Proceedings of the 2020 Conference on Fairness,
  Accountability, and Transparency}, pp.\  648--657, 2020{\natexlab{b}}.

\bibitem[Breiman(2001)]{breiman2001}
Breiman, L.
\newblock Statistical modeling: The two cultures (with comments and a rejoinder
  by the author).
\newblock \emph{Statist. Sci.}, 16\penalty0 (3):\penalty0 199--231, 2001.

\bibitem[Canini et~al.(2016)Canini, Cotter, Gupta, Fard, and
  Pfeifer]{gupta2016lattices}
Canini, K., Cotter, A., Gupta, M.~R., Fard, M.~M., and Pfeifer, J.
\newblock Fast and flexible monotonic functions with ensembles of lattices.
\newblock In \emph{Proceedings of the 30th International Conference on Neural
  Information Processing Systems}, NIPS’16, pp.\  2927–2935, Red Hook, NY,
  USA, 2016. Curran Associates Inc.
\newblock ISBN 9781510838819.

\bibitem[Chen et~al.(2018)Chen, Li, Tao, Barnett, Su, and Rudin]{chen2018looks}
Chen, C., Li, O., Tao, C., Barnett, A.~J., Su, J., and Rudin, C.
\newblock This looks like that: Deep learning for interpretable image
  recognition, 2018.

\bibitem[Dhurandhar et~al.(2018)Dhurandhar, Chen, Luss, Tu, Ting, Shanmugam,
  and Das]{dhur2018explanations}
Dhurandhar, A., Chen, P.-Y., Luss, R., Tu, C.-C., Ting, P., Shanmugam, K., and
  Das, P.
\newblock Explanations based on the missing: Towards contrastive explanations
  with pertinent negatives, 2018.

\bibitem[Genest et~al.(2009)Genest, Gendron, and
  Bourdeau-Brien]{genest2009advent}
Genest, C., Gendron, M., and Bourdeau-Brien, M.
\newblock The advent of copulas in finance.
\newblock \emph{The European journal of finance}, 15\penalty0 (7-8):\penalty0
  609--618, 2009.

\bibitem[Gonçalves et~al.(2020)Gonçalves, Ray, Soper, Stevens, Coyle, and
  Sales]{goncalves2020medical}
Gonçalves, A.~R., Ray, P., Soper, B., Stevens, J.~L., Coyle, L., and Sales,
  A.~P.
\newblock Generation and evaluation of synthetic patient data.
\newblock \emph{BMC Medical Research Methodology}, 20:\penalty0 1471--2288,
  2020.
\newblock \doi{10.1186/s12874-020-00977-1}.

\bibitem[Grath et~al.(2018)Grath, Costabello, Van, Sweeney, Kamiab, Shen, and
  Lecue]{grath2018interpretable}
Grath, R.~M., Costabello, L., Van, C.~L., Sweeney, P., Kamiab, F., Shen, Z.,
  and Lecue, F.
\newblock Interpretable credit application predictions with counterfactual
  explanations, 2018.

\bibitem[Guyon(2003)]{guyonTechReport}
Guyon, I.
\newblock {Design of experiments of the NIPS 2003 variable selection
  benchmark}.
\newblock Technical report, ClopiNet, 01 2003.
\newblock URL
  \url{http://clopinet.com/isabelle/Projects/NIPS2003/Slides/NIPS2003-Datasets.pdf}.

\bibitem[Howe et~al.(2017)Howe, Stoyanovich, Ping, Herman, and
  Gee]{howe2017synthetic}
Howe, B., Stoyanovich, J., Ping, H., Herman, B., and Gee, M.
\newblock Synthetic data for social good.
\newblock \emph{CoRR}, abs/1710.08874, 2017.
\newblock URL \url{http://arxiv.org/abs/1710.08874}.

\bibitem[Ibrahim et~al.(2019)Ibrahim, Louie, Modarres, and
  Paisley]{ibrahim2019gam}
Ibrahim, M., Louie, M., Modarres, C., and Paisley, J.~W.
\newblock Global explanations of neural networks: Mapping the landscape of
  predictions.
\newblock In Conitzer, V., Hadfield, G.~K., and Vallor, S. (eds.),
  \emph{Proceedings of the 2019 {AAAI/ACM} Conference on AI, Ethics, and
  Society, {AIES} 2019, Honolulu, HI, USA, January 27-28, 2019}, pp.\
  279--287. {ACM}, 2019.
\newblock \doi{10.1145/3306618.3314230}.
\newblock URL \url{https://doi.org/10.1145/3306618.3314230}.

\bibitem[Jaworski et~al.(2013)Jaworski, Durante, and Wolfgang]{jaworski_2013}
Jaworski, P., Durante, F., and Wolfgang, H.
\newblock \emph{Copulae in mathematical and quantitative finance: proceedings
  of the workshop held in Cracow, 10-11 July 2012}.
\newblock Springer, 2013.

\bibitem[Kim et~al.(2015)Kim, Shah, and Doshi-Velez]{kim2015mind}
Kim, B., Shah, J.~A., and Doshi-Velez, F.
\newblock Mind the gap: A generative approach to interpretable feature
  selection and extraction.
\newblock In \emph{Advances in Neural Information Processing Systems}, pp.\
  2260--2268, 2015.

\bibitem[Klaise et~al.(2020)Klaise, Van~Looveren, Vacanti, and Coca]{alibi}
Klaise, J., Van~Looveren, A., Vacanti, G., and Coca, A.
\newblock \emph{Alibi: Algorithms for monitoring and explaining machine
  learning models}, 2020.
\newblock URL \url{https://github.com/SeldonIO/alibi}.

\bibitem[Krakovna \& Doshi-Velez(2016)Krakovna and
  Doshi-Velez]{krakovna2016increasing}
Krakovna, V. and Doshi-Velez, F.
\newblock Increasing the interpretability of recurrent neural networks using
  hidden markov models, 2016.

\bibitem[Lee et~al.(2019)Lee, Alvarez{-}Melis, and
  Jaakkola]{lee2019locallylinear}
Lee, G., Alvarez{-}Melis, D., and Jaakkola, T.~S.
\newblock Towards robust, locally linear deep networks.
\newblock \emph{CoRR}, abs/1907.03207, 2019.
\newblock URL \url{http://arxiv.org/abs/1907.03207}.

\bibitem[Lundberg \& Lee(2017)Lundberg and
  Lee]{lundbergUnifiedApproachInterpreting}
Lundberg, S.~M. and Lee, S.-I.
\newblock A unified approach to interpreting model predictions.
\newblock In Guyon, I., Luxburg, U.~V., Bengio, S., Wallach, H., Fergus, R.,
  Vishwanathan, S., and Garnett, R. (eds.), \emph{Advances in Neural
  Information Processing Systems 30}, pp.\  4765--4774. Curran Associates,
  Inc., 2017.
\newblock URL \url{https://arxiv.org/pdf/1705.07874.pdf}.

\bibitem[Nelsen(1999)]{nelsen_1999}
Nelsen, R.~B.
\newblock \emph{An introduction to copulas}.
\newblock Springer, 1999.

\bibitem[Pedregosa et~al.(2011)Pedregosa, Varoquaux, Gramfort, Michel, Thirion,
  Grisel, Blondel, Prettenhofer, Weiss, Dubourg, Vanderplas, Passos,
  Cournapeau, Brucher, Perrot, and Duchesnay]{scikit-learn}
Pedregosa, F., Varoquaux, G., Gramfort, A., Michel, V., Thirion, B., Grisel,
  O., Blondel, M., Prettenhofer, P., Weiss, R., Dubourg, V., Vanderplas, J.,
  Passos, A., Cournapeau, D., Brucher, M., Perrot, M., and Duchesnay, E.
\newblock Scikit-learn: Machine learning in {P}ython.
\newblock \emph{Journal of Machine Learning Research}, 12:\penalty0 2825--2830,
  2011.

\bibitem[Ping et~al.(2017)Ping, Stoyanovich, and Howe]{ping2017datasynthesizer}
Ping, H., Stoyanovich, J., and Howe, B.
\newblock Datasynthesizer: Privacy-preserving synthetic datasets.
\newblock In \emph{Proceedings of the 29th International Conference on
  Scientific and Statistical Database Management}, pp.\  42:1--42:5, 2017.

\bibitem[Ribeiro et~al.(2016)Ribeiro, Singh, and Guestrin]{ribiero2016lime}
Ribeiro, M.~T., Singh, S., and Guestrin, C.
\newblock "why should {I} trust you?": Explaining the predictions of any
  classifier.
\newblock \emph{CoRR}, abs/1602.04938, 2016.
\newblock URL \url{http://arxiv.org/abs/1602.04938}.

\bibitem[Salmon(2012)]{salmon2012formula}
Salmon, F.
\newblock The formula that killed wall street.
\newblock \emph{Significance}, 9\penalty0 (1):\penalty0 16--20, 2012.

\bibitem[TeamHG-Memex(2019)]{eli5}
TeamHG-Memex.
\newblock \emph{Welcome to ELI5’s documentation!}
\newblock TeamHG-Memex, 2019.
\newblock URL \url{https://eli5.readthedocs.io/en/latest/#}.

\bibitem[van~der Linden et~al.(2019)van~der Linden, Haned, and
  Kanoulas]{linden2019global}
van~der Linden, I., Haned, H., and Kanoulas, E.
\newblock Global aggregations of local explanations for black box models, 2019.

\bibitem[Wachter et~al.(2017)Wachter, Mittelstadt, and
  Russell]{wachter2017counterfactual}
Wachter, S., Mittelstadt, B., and Russell, C.
\newblock Counterfactual explanations without opening the black box: Automated
  decisions and the gdpr, 2017.

\bibitem[Yang \& Kim(2019)Yang and Kim]{yangBenchmarkingAttributionMethods2019}
Yang, M. and Kim, B.
\newblock Benchmarking {{Attribution Methods}} with {{Relative Feature
  Importance}}.
\newblock \emph{CoRR}, abs/1907.09701, 2019.
\newblock URL \url{http://arxiv.org/abs/1907.09701}.

\end{thebibliography}
\bibliographystyle{icml2020}

\end{document}